\documentclass{article}

\usepackage{times}
\usepackage{graphicx} % more modern

\usepackage{subfigure} 

\usepackage{natbib}

\usepackage{algorithm}
\usepackage{algorithmic}

\usepackage[accepted]{icml2015_hack}

\usepackage{amsmath}
\usepackage{amssymb}
\graphicspath{{./},{./figs/}}
\DeclareGraphicsExtensions{.pdf,.jpeg,.png}

\newcommand*\diff{\mathop{}\!\mathrm{d}}

\renewcommand{\vec}[1]{\mathbf{#1}}

\icmltitlerunning{Spike and Slab Gaussian Process Latent Variable Models}

\begin{document} 

\twocolumn[
\icmltitle{Spike and Slab Gaussian Process Latent Variable Models}

% unless you've provided the [accepted] option to the icml2015
\icmlauthor{Zhenwen Dai}{z.dai@sheffield.ac.uk}
\icmladdress{Department of Computer Science,
            University of Sheffield, UK}
\icmlauthor{James Hensman}{james.hensman@sheffield.ac.uk}
\icmladdress{Department of Computer Science,
            University of Sheffield, UK}
\icmlauthor{Neil Lawrence}{n.lawrence@sheffield.ac.uk}
\icmladdress{Department of Computer Science,
            University of Sheffield, UK}

\icmlkeywords{Gaussian Process, GP-LVM, spike and slab, variational inference}

\vskip 0.3in
]

\begin{abstract} 
The Gaussian process latent variable model (GP-LVM) is a popular
approach to non-linear probabilistic dimensionality reduction. One
design choice for the model is the number of latent variables.  We
present a spike and slab prior for the GP-LVM and propose an efficient
variational inference procedure that gives a lower bound of the log
marginal likelihood. The new model provides a more principled approach
for selecting latent dimensions than the standard way of thresholding
the length-scale parameters. The effectiveness of our approach is
demonstrated through experiments on real and simulated data. Further,
we extend multi-view Gaussian processes that rely on sharing latent
dimensions (known as manifold relevance determination) with spike and
slab priors. This allows a more principled approach for selecting a
subset of the latent space for each view of data. The extended model
outperforms the previous state-of-the-art when applied to a
cross-modal multimedia retrieval task.
\end{abstract}

\section{Introduction}

% Show the advantages of gplvm as nonlinear dimension reduction.

Gaussian Process latent variable models (GP-LVM) reduce the
dimensionality of data by establishing a mapping from a low
dimensional latent space, $X$, to a high dimensional observed space,
$Y$, through Gaussian Process (GP) \citep{Lawrence:pnpca05, TitsiasLawrence2010}. The
nonparametric nature of GP and the flexibility of using non-linear
kernels enables GP-LVM to produce compact representations of
data. GP-LVM has been successfully applied to various domains as a
dimension reduction method. For example, \citet{BuettnerTheis2012}
used GP-LVM for resolving differences in single-cell gene expression
patterns from zygote to blastocyst, and \citet{LuTang2014} developed a
discriminative GP-LVM for face verification that was the first
to surpass human-level performance.

%One of the major improvements for GP-LVM is called Bayesian GP-LVM  \citep{Titsias:bayesGP-LVM10}, where a unity Gaussian prior is introduced for the latent variable $X$, and $X$ is analytically maginalised out. In general, such marginalisation is intractable, but Bayesian GP-LVM achieves it through a variational approximation. The marginalisation of $X$ allows us to consider uncertainty in latent space, and its sparse GP formulation enables its application to much bigger dataset.

When applying GP-LVM to dimension reduction problems, a key parameter
of choice is the dimensionality of the latent space $Q$. A larger
latent dimensionality can correspond to a significantly higher number
of parameters, which potentially leads to overfitting. A standard
approach for choosing the latent dimensionality of GP-LVM is to look
at the values of the length-scale parameters of kernel
functions. These parameters characterise the scales of individual
latent dimensions. The underlying latent function can only vary
``slowly'' along the latent dimension with a high length-scale. When
the length scale of a latent dimension is significantly larger than
the ones of other dimensions, the influence of this latent dimension
to overall (co)variances is negligible. Therefore, the number of
latent dimensions is conventionally determined according to the
length-scale parameters, typically by comparing to a manually chosen
threshold. It closely relates to the idea of automatic relevance
determination (ARD) regression \citep{Rasmussen:2005:GPM:1162254},
where the ARD parameters are defined as one over the length scales.  A
limitation of this approach is that the choice of threshold could
involve a lot of hand tuning. Furthermore, for non-linear kernels like
the exponentiated quadratic form, it is difficult to figure out
whether a relatively high length scale means that the latent dimension
is ``slowly'' switched off or the latent dimension is very smooth
\cite{Vehtari2001}.

A more principled approach is preferred for§ automating the selection
of the number of latent dimensions. It means to let the model decide
which latent dimensions it really needs.  Driven by this idea, we
introduce spike and slab prior \citep{MitchellBeauchamp1988,
  citeulike:4523099, West03bayesianfactor} for the latent variable
$X$. A spike and slab prior contains binary variables, which allows
the model to probabilistically discard latent dimensions.  Given
observed data $Y$, the posterior probability of a latent dimension
being used can be derived from the model definition. It offers a
principled approach for selecting latent dimensions. However, the
exact inference of the posterior distribution of latent variables is
intractable, because there is no closed form solution for the integral
of latent variables in the log marginal likelihood. In this paper, we
derive a closed form variational lower bound of the log marginal
likelihood for the spike and slab GP-LVM and develop an efficient inference
method for posterior distributions of latent representations $X$ and
switching variables.
%Mention the variational approach that we take.
In spike and slab models standard mean field approximations are problematic due to the strong correlation between switch variables and input variables. Our variational approach assumes a conditional dependence between input variables and switch variables. It is closely related to the ideas like structured mean field \cite{Saul95exploitingtractable,Xing:2002:GMF:2100584.2100655}.
This allows us to efficiently infer the latent representations of data while simultaneously determining the active latent dimensions.

In the literature, the spike and slab prior has been used for variable
selection in various regression models. For instance,
\citet{CarbonettoStephens2012} developed a variational inference
method with spike and slab prior for Bayesian variable selection in
linear models. \citet{SavitskyEtAl2011} introduced spike and slab
prior to length scale parameters in variable selection of GP
regression models. They proposed inference through a Markov chain
Monte Carlo (MCMC) scheme. In this paper, both switch variables and
input variables $X$ are \emph{variationally} integrated out. This
enables us to infer latent representations for dimension
reduction. Note this contrasts with regression, where available inputs
are only switched on or off. Here we are selecting both the number of
available inputs and their nature (through the latent variable
approach). Spike and slab priors have also been used in unsupervised
learning for sparse \emph{linear} models (i.e. sparse coding) with
variational or truncated approximations \cite{TitsiasLazaroGredilla2011,
  SheikhEtAl2014}. The spike and slab GP-LVM we introduced is much more
flexible because it allows for the encoding of \emph{non-linear}
relationships through appropriate Gaussian process covariance
selection.

Through principled formulation of the selection of latent dimensions,
our efficient variational approach allows us to extend the
\emph{multi-view} learning of GP with explicit separation of latent
spaces for related views of a data set. The multi-view GP model, known
as manifold relevance determination (MRD) in
\cite{Damianou:manifold12}, develops latent spaces that are shared
amongst the different views and latent spaces that are particular to
each given view. It formulation can be distilled as a set of
inter-related GP-LVM models which share latent dimensions. Learning in
the model consists of assigning each GP-LVM to a separate sub set of
the latent dimensions through adjustment of length-scale
parameters. Therefore, with applying a GP-LVM to each view of data,
each view can ``softly'' decide which latent dimensions to use by
variation of the length-scale parameter. However, this introduces the
same ambiguity we referred to above. A threshold must be selected for
deciding when a latent dimension is being ignored. With the spike and
slab prior, different views can focus on a subset of the latent space
by discrete switching of the unnecessary dimensions. This provides a
more principled approach for multi-view learning with GPs. Our
variational spike and slab GP-LVM is easily extended to handle this
particular case.

Before introducing the new variational approximation, we first review
the GP-LVM and its Bayesian counterpart. We then introduce our spike
and slab GP-LVM, and the extension to MRD. Finally we empirically
demonstrate the effectiveness of our model in selecting latent
dimensions with both synthetic and real data. We demonstrate the new
multi-view approach on an image-text dataset, in which gives
significantly better results than the previous state-of-the-art.

\section{Gaussian Process Latent Variable Model}

For unsupervised learning, we typically assume a set of observed data $Y \in \mathbb{R}^{N \times D}$ with $N$ datapoints and $D$ dimensions for each datapoint. Our aim is to obtain a latent representations of the observed data, which we denote by $X \in \mathbb{R}^{N \times Q}$, where $Q$ is the number of latent space. In GP-LVM, the relationship between latent representations and data is given by a Gaussian process 
\begin{align}\label{eqn:f_X}
p(\vec{f}_d | X) = \mathcal{N}(\vec{f}_d; 0, K),
\end{align}
and for simplicity we assume Gaussian noise,
\begin{equation}
p(\vec{y}_d|\vec{f}_d) = \mathcal{N}(\vec{y}_d; \vec{f}_d, \beta^{-1}\mathbb{I}),
\end{equation}
where $\vec{y}_d \in \mathbb{R}^{N} $ is the $d$th dimension of the observed data, $\vec{f}_d$ is called the noise-free observation, and $K$ is the covariance matrix of $X$ computed according a kernel function $k(x,x')$. By maximizing the log likelihood $\log p(Y|X)$, the point estimates of latent representations $X$ can be obtained \cite{Lawrence:pnpca05}. However, the point estimation of $X$ implies fitting a lot of parameters, therefore, the resulting model is prone to overfitting. \citet{TitsiasLawrence2010} overcome this limitation by introducing a unit Gaussian prior for $X$ and deriving a variational lower bound of the log marginal likelihood (known as Bayesian GP-LVM).
In their model, the sparse GP formulation \cite{Titsias:variational09} is used. This relies on augmenting our variable space with a set of inducing variables, $\vec{u}$, and the model becomes:
\begin{align}
p(\vec{f}_d | \vec{u}_d, X, Z) =& \mathcal{N}(\vec{f}_d; K_{fu} K_{uu}^{-1} \vec{u}_d, \nonumber\\ &\quad K_{ff}-K_{fu} K_{uu}^{-1} K_{fu}^\top),\\
p(\vec{u}_d | Z) =& \mathcal{N} (\vec{u}_d ; 0, K_{uu}),\\
p(X) =& \prod_{n=1}^N \mathcal{N}(\vec{x}_n; 0, \mathbb{I}),
\end{align}
where $\vec{u}_d$ is the inducing variable for $d$th dimension, and
$Z$ is the inducing input. $K_{ff}$ and $K_{uu}$ are the covariance
matrices for $X$ and $Z$ respectively, and $K_{fu}$ is the cross
covariance matrix between $X$ and $Z$. Then, the log marginal
likelihood of the model is defined as
\begin{equation}
p(Y) = \int \prod_{d=1}^{D} p(\vec{y}_d| X) p(X) \diff{X}.
\end{equation}
For tractability, the integral of $X$ in the log marginal likelihood is approximated variationally. A variational posterior distribution of $X$ is defined as:
\begin{equation}
q(X) = \prod_{n=1}^{N} \mathcal{N}(\vec{x}_n; \vec{\mu}_n, S_n).
\end{equation}
With the assumption $p(\vec{f}_d| \vec{y}_d,
\vec{u}_d, X) = p(\vec{f}_d | \vec{u}_d, X)$, a lower bound of
log marginal likelihood can be obtained with Jensen's inequality:
\begin{align}
\log p(Y) \geq &\sum_{d=1}^{D} \tilde{F}_d(q) - \text{KL}(q(X)\|p(X)),\\
\tilde{F}_d(q) =& \log \Big[\frac{(\beta)^{\frac{N}{2}} |K_{uu}|^{\frac{1}{2}}}{(2\pi)^{\frac{N}{2}} | \beta \Psi_2 + K_{uu}|^{\frac{1}{2}}} e^{-\frac{1}{2}y_{d}^\top W y_{d}} \Big] \nonumber\\
&-\frac{\beta\psi_0}{2} + \frac{\beta}{2}\text{Tr}(K_{uu}^{-1}\Psi_2),\label{equ:Ft_d}\\
\text{KL}(q(X)&\|p(X)) = \int q(X) \log \frac{q(X)}{p(X)} dX,
\end{align}
where $W= \beta \mathbb{I} - \beta^2 \Psi_1 (\beta \Psi_2+K_{uu})^{-1}\Psi_1^\top$. $\psi_0$, $\Psi_1$ and $\Psi_2$ are the expectation of the covariance matrices w.r.t.\ $q(X)$, i.e.\ $\psi_0 = \text{Tr}( \mathbb{E}_{q(X)} [K_{ff}])$,  $\Psi_1 = \mathbb{E}_{q(X)} [ K_{fu}]$, $\Psi_2 = \mathbb{E}_{q(X)} [K_{fu}^\top K_{fu} ]$. With this formulation, the latent variable $X$ is variationally integrated out. It leads to a closed-form lower bound of the log marginal likelihood. 

% explain the ARD latent dimension selection
In this formulation, the selection of latent dimensionality relies on parameters called length scales, each of which is applied to scale separately the input dimensions, e.g., $\vec{l}$ in the exponentiated quadratic kernel function $k(\vec{x}
, \vec{x}')= \sigma_f^2 \exp (-\frac{1}{2}\sum_{q=1}^Q (x_q - x'_q)^2 / l_q^2)$. If the length scale of a latent dimension is significantly lower than the other latent dimensions, the influence of this latent dimension to overall (co)variances is negligible. Therefore, a typical approach selects latent dimensions by thresholding their length scales. In this work, we propose a more principled approach, where each latent dimension is intrinsically considered whether it is used or not,  by explicitly introducing a latent switching variable $\vec{b} \in \{0,1\}^{Q}$ that consists of a set of binary variables, each controlling the usage of a latent dimension. 

\section{Spike and Slab GP-LVM}

We introduce a switch variable  $\vec{b} \in \{0,1\}^{Q}$ that determines whether a particular latent dimension is used or not. The way that the switch variable controls the usage of a latent dimension is done by replacing the input variable $\vec{x}_n$ in the original Bayesian GP-LVM  by $\vec{x}_n \circ \vec{b}$, where $\circ$ denotes the element-wise multiplication. If a binary variable in $\vec{b}$ is zero, the input of the corresponding dimension to the underlying GP becomes zero. As the input variable $X$ has a Gaussian prior, the combination of $X$ and $\vec{b}$ is known as a spike and slab prior. The prior distribution of the latent switch variables $\vec{b}$ are goverend by Bernoulli distributions,
\begin{equation}\label{eqn:b_prior}
p(\vec{b})= \prod_{q=1}^Q \pi^{b_{q}}(1-\pi)^{(1-b_{q})},
\end{equation}
where $\pi$ is the prior probability, which typically takes the value $0.5$. 
Due to the introduction of $\vec{b}$, all the $\vec{x}_n$ in original covariance matrices are replaced with $\vec{b} \circ \vec{x}_n$. This changes the form of the cross covariance matrix $K_{fu}$. In the case of the exponentiated quadratic kernel it becomes
\begin{equation}
(K_{fu})_{nm} = \sigma_f^2 \exp \left( -\sum_{q=1}^{Q}\frac{(b_q x_{nq} - z_{mq})^2}{2l_q^2}\right),
\end{equation}
where $(\cdot)_{nm}$ indicates the $n$th row and $m$th column of the matrix, while $K_{uu}$ is not affected.  To make inference tractable, we take a variational approach, where we assume a variational posterior distribution for the spike and slab model. As the switch variable and the slab variable are strongly correlated, the variational posterior is defined as a conditional distribution,
\begin{align}
q(\vec{b}) = \prod_{q=1}^Q \gamma_q^{b_q} (1-\gamma_q)^{(1-b_q)},\\
q(x_{nq}|b_q = 1) = \mathcal{N} (x_{nq};  \mu_{nq}, s_{nq}),
\end{align}
where $\gamma_q$ is the posterior probability of the $q$th dimension being used (the overall probability of the slab part), and $\mu_{nq}$ and $s_{nq}$ are the mean and variance of the variational posterior distribution for the slab part. It is closely related to the ideas like structured mean field \citep{Saul95exploitingtractable,Xing:2002:GMF:2100584.2100655}. Therefore, the lower bound of log marginal likelihood becomes
\begin{align}
\log p(Y) \geq &\sum_{d=1}^{D} \tilde{F}_d(q) - \text{KL}(q(\vec{b}, X)\|p(\vec{b})p(X)),\\
\text{KL}(q(\vec{b}, X)&\|p(\vec{b}) p(X)) =\nonumber\\ 
 &\sum_{\vec{b}} \int q(\vec{b}, X) \log \frac{q(\vec{b}, X)}{p(\vec{b})p(X)} dX,
\end{align}
where $\tilde{F}_d(q)$ keeps the same form as in (\ref{equ:Ft_d}), while $\Psi_1$ and $\Psi_2$ need to adapted according to the new $K_{fu}$. Then, the new $\Psi_1$ and $\Psi_2$ are defined as:
\begin{align}
(\Psi_1)_{nm} =& \sum_{\vec{b}} q(\vec{b})  \int k(\vec{b} \circ \vec{x}_n, \vec{z}_m) q(\vec{x}_n | \vec{b}) d\vec{x}_n,\label{eqn:ssgplvm_psi1}\\
(\Psi_2)_{mm'} =& \sum_{n=1}^N \sum_{\vec{b}} q(\vec{b})  \int k(\vec{b} \circ \vec{x}_n, \vec{z}_m) \nonumber\\
&\quad k(\vec{z}_{m'}, \vec{b} \circ \vec{x}_n) q(\vec{x}_n | \vec{b}) d\vec{x}_n. \label{eqn:ssgplvm_psi2}
\end{align}
For some kernels, the closed-form solution for $\psi_0$, $\Psi_1$ and $\Psi_2$ can be obtained. For example, $\psi_0$, $\Psi_1$ and $\Psi_2$ of the exponentiated quadratic kernel are derived as:
\begin{align}
\psi_0 &= N \sigma_f^2, \label{eqn:psi0}\\
(\Psi_1)_{nm}  &=  \sigma_f^2 \prod_{q=1}^Q \Big[\frac{\gamma_{nq}}{(s_{nq}/l_q+1)^{\frac{1}{2}}} e^{-\frac{1}{2}\frac{(\mu_{nq}-z_{mq})^2}{ s_{nq}+l_q}} \nonumber\\
&\hspace{17mm} + (1-\gamma_{nq}) e^{-\frac{z_{mq}^2}{2l_q}}\Big],\label{eqn:psi1}\\
(\Psi_2)_{mm'} &=  \sigma_f^4\sum_{n=1}^N \prod_{q=1}^Q \Big[\frac{\gamma_{nq}}{(2s_{nq}/l_{q}+1)^{\frac{1}{2}}} \nonumber \\
&\hspace{10mm}e^{-\frac{(z_{mq}-z_{m'q})^2}{4l_q}-\frac{(\mu_{nq}-(z_{mq}+z_{m'q})/2)^2}{2 s_{nq}+l_q}} \nonumber\\
&\hspace{10mm}+ (1-\gamma_{nq}) e^{-\frac{(z_{mq}^2+z_{m'q}^2)}{2l_q}} \Big].\label{eqn:psi2}
\end{align}

Note that the distribution $q(x_{nq} | b_q = 0)$, because, as the switch variable is zero, the slab variable does not influence the likelihood anymore, so that $q(x_{nq} | b_q = 0)$ will only appear inside the KL divergence, which make it always equal to the prior distribution $p(X)$.

%\section{Variational Approximation for Spike and Slab distribution}
%
%\section{Prediction}
%
%\newpage
%\begin{equation}
%\phi(\vec{u}_d) = \mathcal{N} (\vec{u}_d ; \vec{w}_d, \Sigma)
%\end{equation}
%\begin{align}
%L = -\frac{N}{2}\log2\pi -\frac{1}{2}\log |R|  -\frac{1}{2}\vec{y}_d^\top R^{-1} \vec{y}_d -\frac{1}{2}\vec{w}_d^\top (K_{uu}^{-1} \Psi_2 K_{uu}^{-1}+ K_{uu}^{-1}) \vec{w}_d  -\frac{1}{2} \text{Tr}((K_{uu}^{-1} \Psi_2 K_{uu}^{-1}+ K_{uu}^{-1}) \Sigma) \\+ \vec{y}_d^\top \Psi_1 K_{uu}^{-1}\vec{w}_d + \log \frac{|2 \pi \Sigma|^{\frac{1}{2}}}{|2 \pi K_{uu}|^{\frac{1}{2}}}
%- \frac{1}{2} \psi_0  +\frac{1}{2} \text{Tr}(K_{MM}^{-1} \Psi_2)
%\end{align}
%
%The gradients:
%\begin{align}
%\frac{\partial L}{\partial \Psi_2} = 
%\end{align}

\section{Spike and Slab MRD}

\begin{figure}[t]
        \centering

	\includegraphics[width=0.8\linewidth]{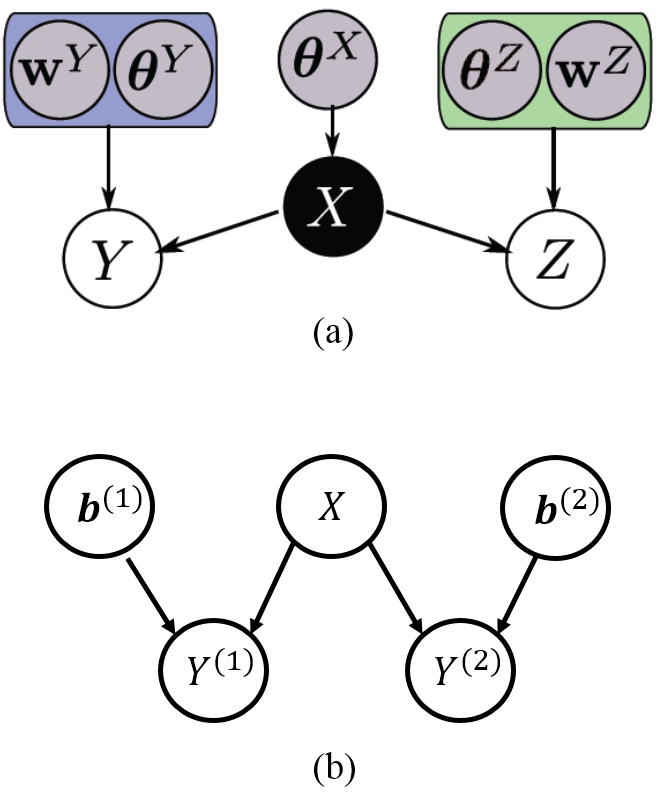}

 \caption{(a) The graphical model for the original MRD, where $Y$ and $Z$ denotes two views of the data and $X$ denote the latent variable. (b) The graphical model for the spike and slab MRD.}\label{fig:mrd}
\end{figure}

In manifold relevance determination \citep{Damianou:manifold12}, multiple views of data are considered simultaneously. The model assumes that those views share some aspects and retain some aspects that are particular to each view. Each latent dimension can both relate to other views in some shared latent dimensions while keeping some private latent dimension for its own (see Fig.\ \ref{fig:mrd}a). In the original MRD paper this effect was achieved through appropriate sharing of ARD parameters within each view. This leads to a soft sharing approach where a particular latent dimension can be used to a greater or lesser extent by each of the views. Spike and slab MRD allows for probabilistic selection of binary variables to perform the sharing (see Fig.\ \ref{fig:mrd}b).

We wish to relate $C$ views $Y^{(c)} \in \mathbb{R}^{N \times D_c}$ of a dataset in our model. We assume the dataset can be represented as a latent variable $X \in \mathbb{R}^{N \times Q}$, in which a different subset of the $Q$ latent dimensions are used to represent each view. The selection of latent dimensions for $c$th view is done by a vector of latent binary variable $\vec{b}_c \in \mathbb{R}^{Q}$ with the prior distribution in (\ref{eqn:b_prior}).
As mentioned in previous section, each view can be modeled by a spike and slab GP:
\begin{equation}
p(Y | X, B) = \prod_{c=1}^{C} p(Y^{(c)} | X, B).
\end{equation} 
With the prior distribution of $X$ and $B$, the marginal distribution of our MRD model is
\begin{equation}
p(Y) = \int \prod_{c=1}^C \sum_{B} p(B) \prod_{d=1}^{D} p(\vec{y}_d^{(c)}| X, B) p(X) \diff{X}
\end{equation}
% The complicated new variational distribution
For a tractable inference algorithm, we wish to introduce a variational approximation for the latent variable $X$ and $B$. Similar to spike and slab GP-LVM, the latent variable $X$ and $B$ are closely correlated, so that we define a conditional variational distribution $q(X,B) = q(X|B) \prod_{c=1}^C q(\vec{b}_c)$, in which there is a variational posterior distribution for each view representing the subset of latent dimensions used by that view, and the posterior distribution for the latent representation of data, which is the same for all the views. In order to be consistent with the choices $B$, $q(X|B)$ is defined as
\begin{align}
q(\vec{b}_c) = \prod_{q=1}^Q \gamma_{cq}^{b_{cq}} (1-\gamma_{cq})^{(1-b_{cq})}, \\ 
q(x_{nq}| \bigvee_{c=1}^C (b_{cq}=1)) = \mathcal{N}(x_{nq}; \mu_{nq}, s_{nq}),
\end{align}
where $\bigvee$ is the \textit{or} operation for binary variables. We denote the conditional variational posterior distribution $q(x_{nq}| \bigvee_{c=1}^C (b_{cq}=1))$ as $q_c(x_{nq})$.  It gives the variational posterior of $X$ for all the views if any of the views decide to use the latent dimension, otherwise the posterior distribution will fall back to the prior, therefore the posterior $q(x_{nq}| \bigwedge_{c=1}^C (b_{cq}=0))$ does not need to be explicitly defined. 
By variationally integrating out the latent variable $X$ and $B$, we obtain a lower bound for our MRD model:
\begin{align}
p(Y) = \sum_{B} p(B) \int  \prod_{c=1}^{C} p(Y^{(c)} | X, \vec{b}_c) p(X|B) \diff{X}
\end{align}
\begin{align}
\log p(Y) \geq & \sum_{c=1}^C\sum_{d=1}^{D} \tilde{F}_d^{(c)}(q) \nonumber\\
&\hspace{6mm}- \text{KL}(q(B, X)\|p(B)p(X)),\\
\text{KL}(q(B, X)&\|p(B) p(X)) = \sum_{B} \int q(B, X) \nonumber\\
& \hspace{2cm} \log \frac{q(B, X)}{p(B)p(X)} dX,
\end{align}

With this definition, we will have a new $\Psi_1$ and $\Psi_2$ correspondingly.
\begin{align}
(\Psi_1^{(c)})_{nm} =& \sum_{B} \prod_{c'=1}^C q(\vec{b}_{c'})  \int k(\vec{b}_c \circ \vec{x}_n, \vec{z}_m) \nonumber\\
&\hspace{28mm} q(\vec{x}_n | B) d\vec{x}_n,\\
(\Psi_2^{(c)})_{mm'} =& \sum_{n=1}^N \sum_{B} \prod_{c'=1}^C q(\vec{b}_{c'})  \int k(\vec{b}_c \circ \vec{x}_n, \vec{z}_m) \nonumber\\
&\quad k(\vec{z}_{m'}, \vec{b}_c \circ \vec{x}_n) q(\vec{x}_n | B) d\vec{x}_n,
\end{align}
however, they lead to exactly the same formulas as (\ref{eqn:psi0}), (\ref{eqn:psi1}) and (\ref{eqn:psi2}). For efficient implementation, the computation of $\psi_0$, $\Psi_1$ and $\Psi_2$ which is usually the bottleneck can be easily parallelized by dividing data points into small groups and evaluating the results in a distributed way \citep{1410.4984,Gal:distributed14}.

%The KL-divergence is different.
%\begin{align}
%\text{KL}&(q(B, X)\|p(B) p(X)) = \sum_{c=1}^C \sum_{q=1}^Q \sum_{b_{cq}} q(b_{cq}) \log \frac{q(b_{cq})}{p(b_{cq})} \nonumber \\
%& +\sum_{q=1}^Q q(b_{cq}=1) \sum_{n=1}^N \int \mathcal{N}(\mu_{nq},s_{nq}) \log \frac{\mathcal{N}(\mu_{nq},s_{nq})}{\mathcal{N}(0,1)}
%\end{align}

%The principled mechanism of disabling latent dimensions 

%Gaussian Process latent variable models have been extend to explore an efficient latent representation from multiple views of data. The latent space is factorized into a common subspace that represents the shared inform across all the views plus a private subspace for each view that keeps the information special about that view. In the previous approach, the MRD is designed as a combination of multiple Bayesian GP-LVMs (one for each view of data), where the latent representations $q(X)$ are shared across all the views. The separation of latent space for each view is archived by letting each view has its own set of ARD parameters 

\section{Experiments}

\begin{figure}[t]
        \centering
        \subfigure[]{\label{fig:syth_datasource}\includegraphics[width=\linewidth]{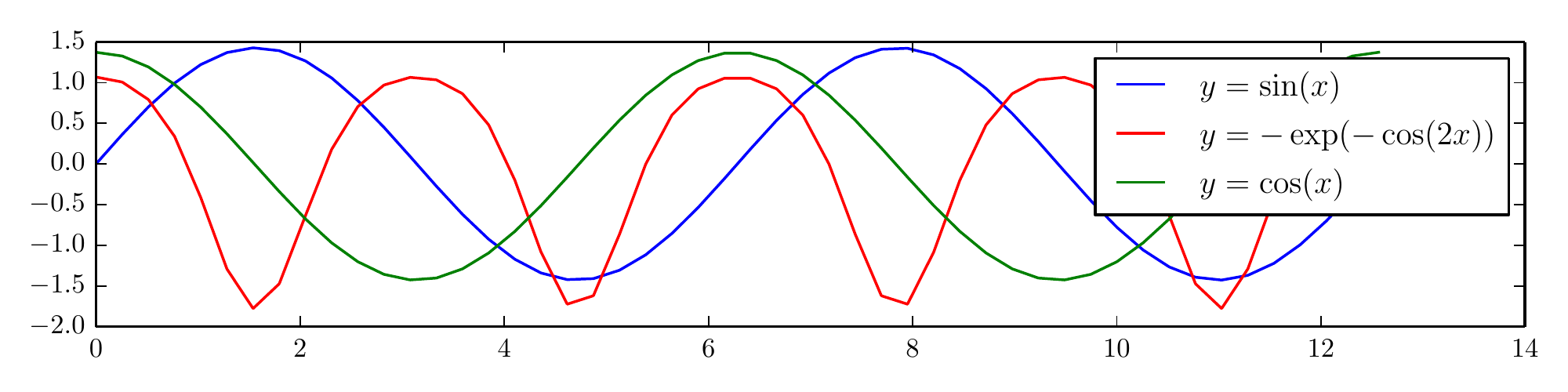}}\\
	\subfigure[]{\label{fig:syth_data}\includegraphics[width=\linewidth]{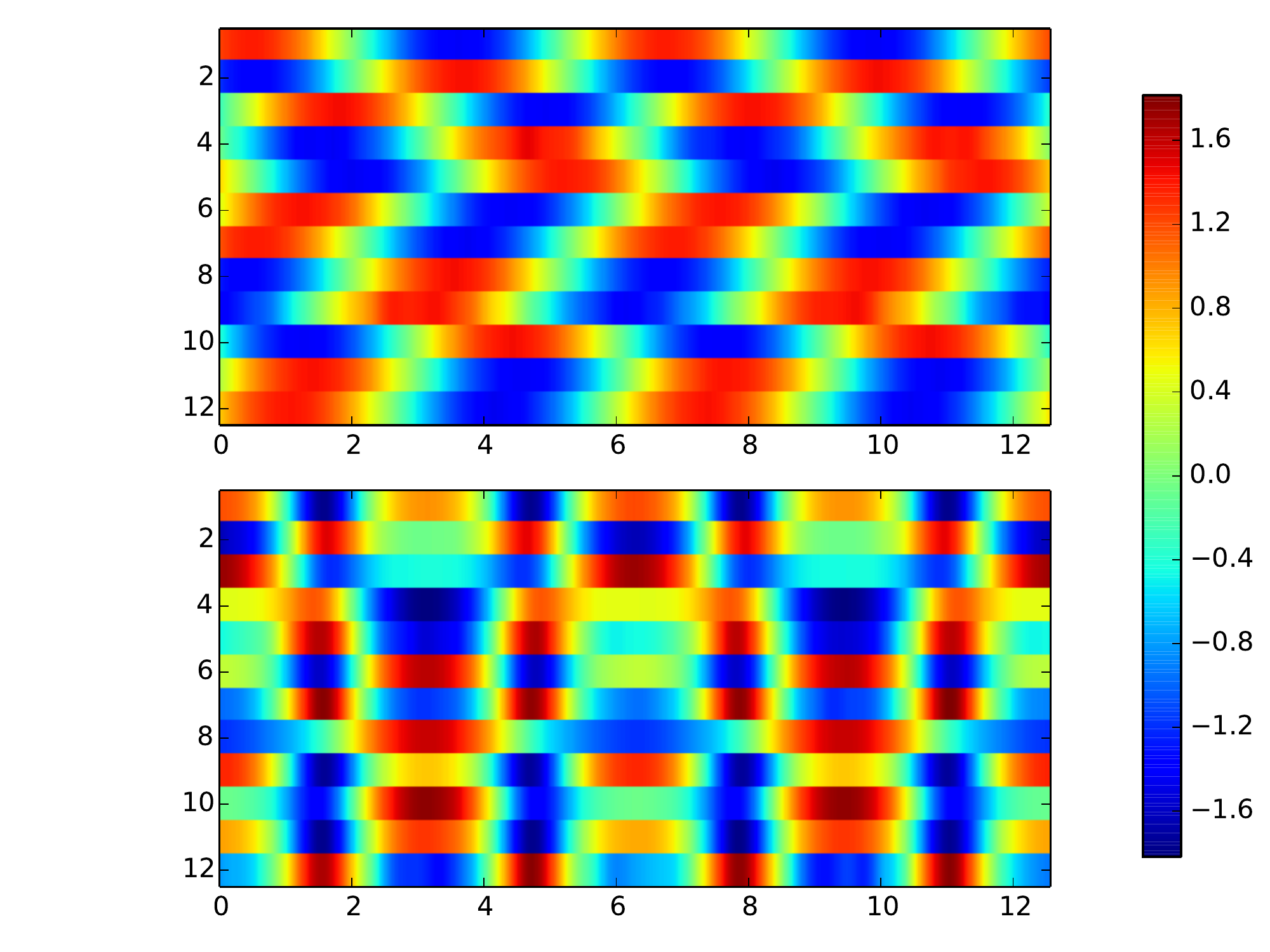}}
 \caption{(a) The three latent signals used for generating observed data. (b) Two observed data generated from the latent signals. The $y$-axis shows the different dimensions in the observed data, $12$ dimensions for each. The $x$-axis shows the data in each dimensions, and $50$ samples are drawn evenly.}
\end{figure}

We first demonstrate our model with synthetic data. We aim to recover
the latent signal from two sets of multi-dimensional observed
data. Then, we show the effectiveness of the switch variable as a
cue for choosing latent dimensions. After that, we apply our SSMRD
model to a text-image dataset, where our model gives significantly
better results than state of the art performance.

\begin{figure}[t]
        \centering
        \subfigure[]{\includegraphics[width=\linewidth]{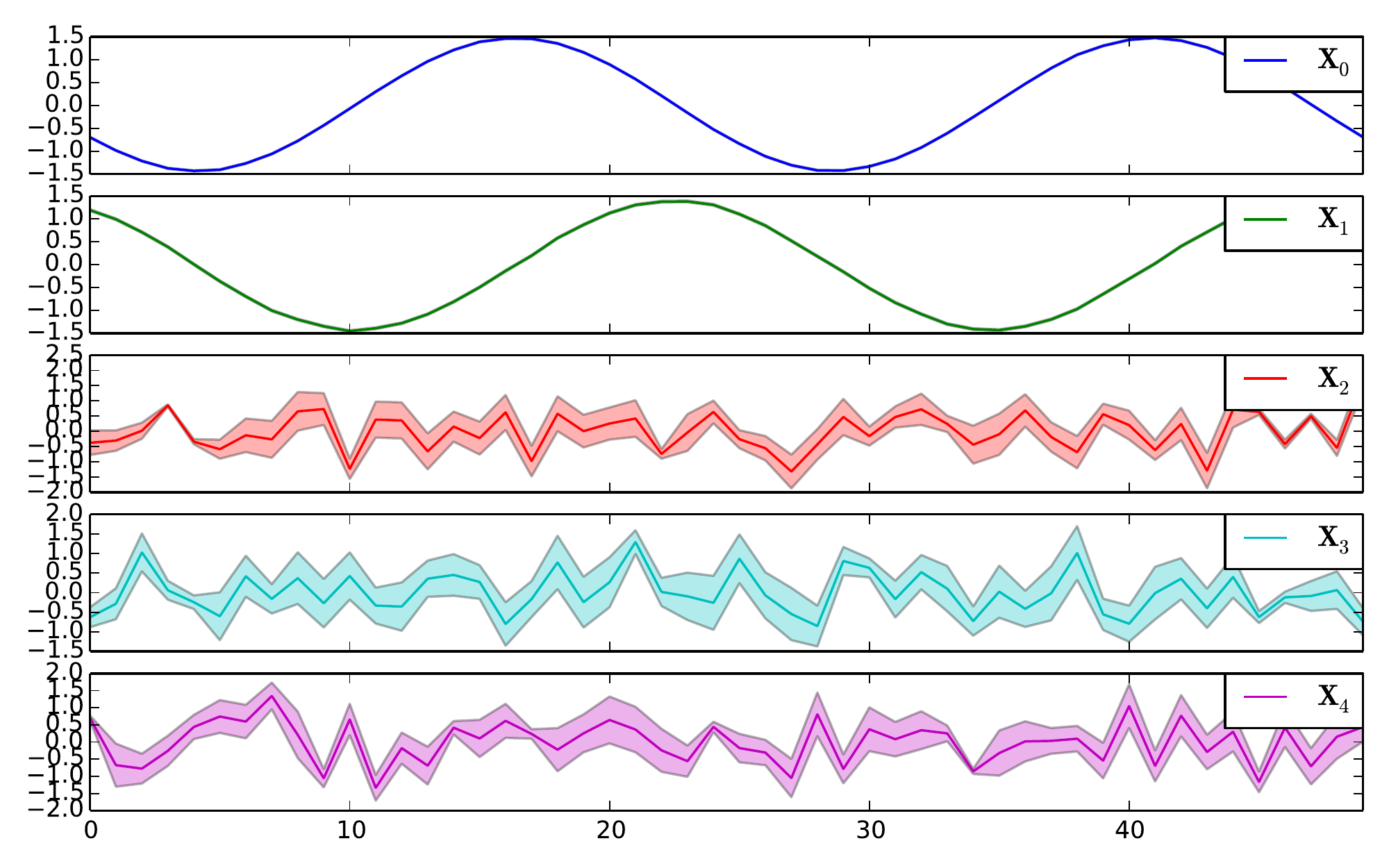}
        	\label{fig:syth_ssgplvmX}}\\
	\subfigure[]{\includegraphics[width=0.48\linewidth]{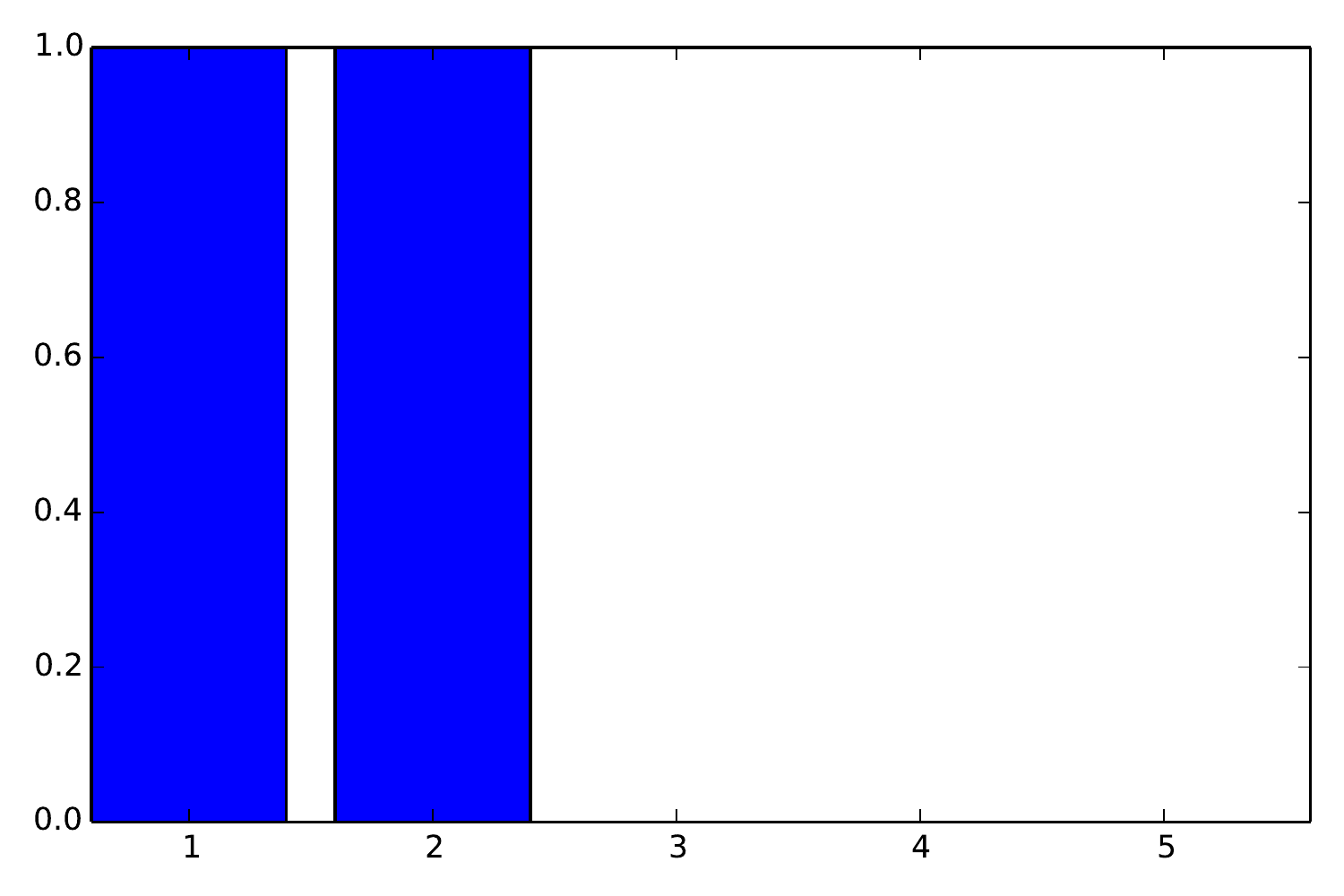}
	\label{fig:syth_ssgplvmXgamma}}
	\subfigure[]{\includegraphics[width=0.48\linewidth]{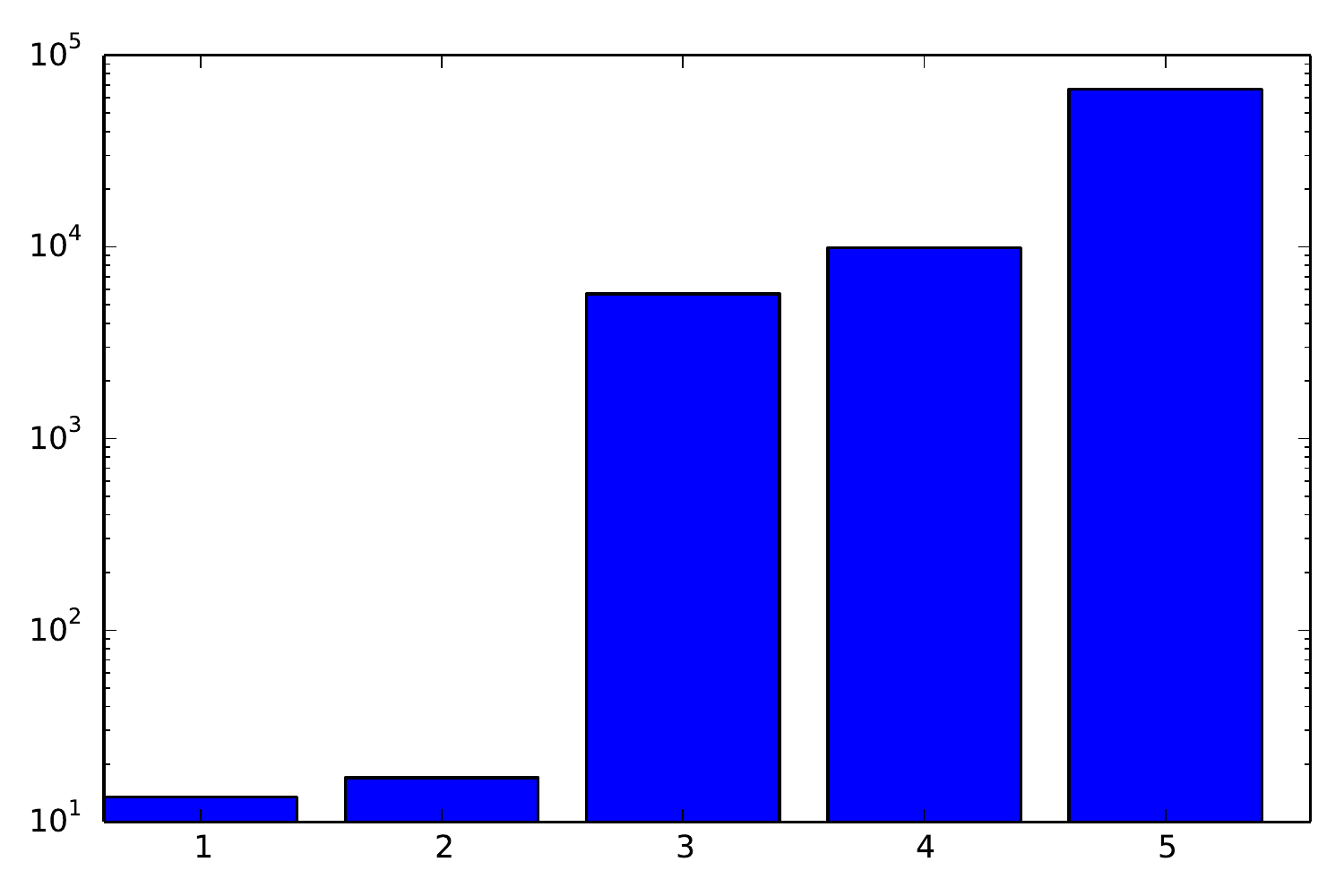}
	\label{fig:syth_ssgplvmlengthscale}}

     \caption{(a) The recovered latent signals according to our SSGP-LVM model from the first signal in Fig.\ \ref{fig:syth_data}. It shows the conditional variational posterior distribution $q_c(X)$. Each row corresponds to a latent dimension, and the curve shows the mean, and the width of the colored region shows the variance.
 (b) The learned posterior probability for the switch variable $\vec{b}$. (c) The learned length scales of latent dimensions.}
\end{figure}

\begin{figure}[t]
        \centering
        \subfigure[]{\includegraphics[width=\linewidth]{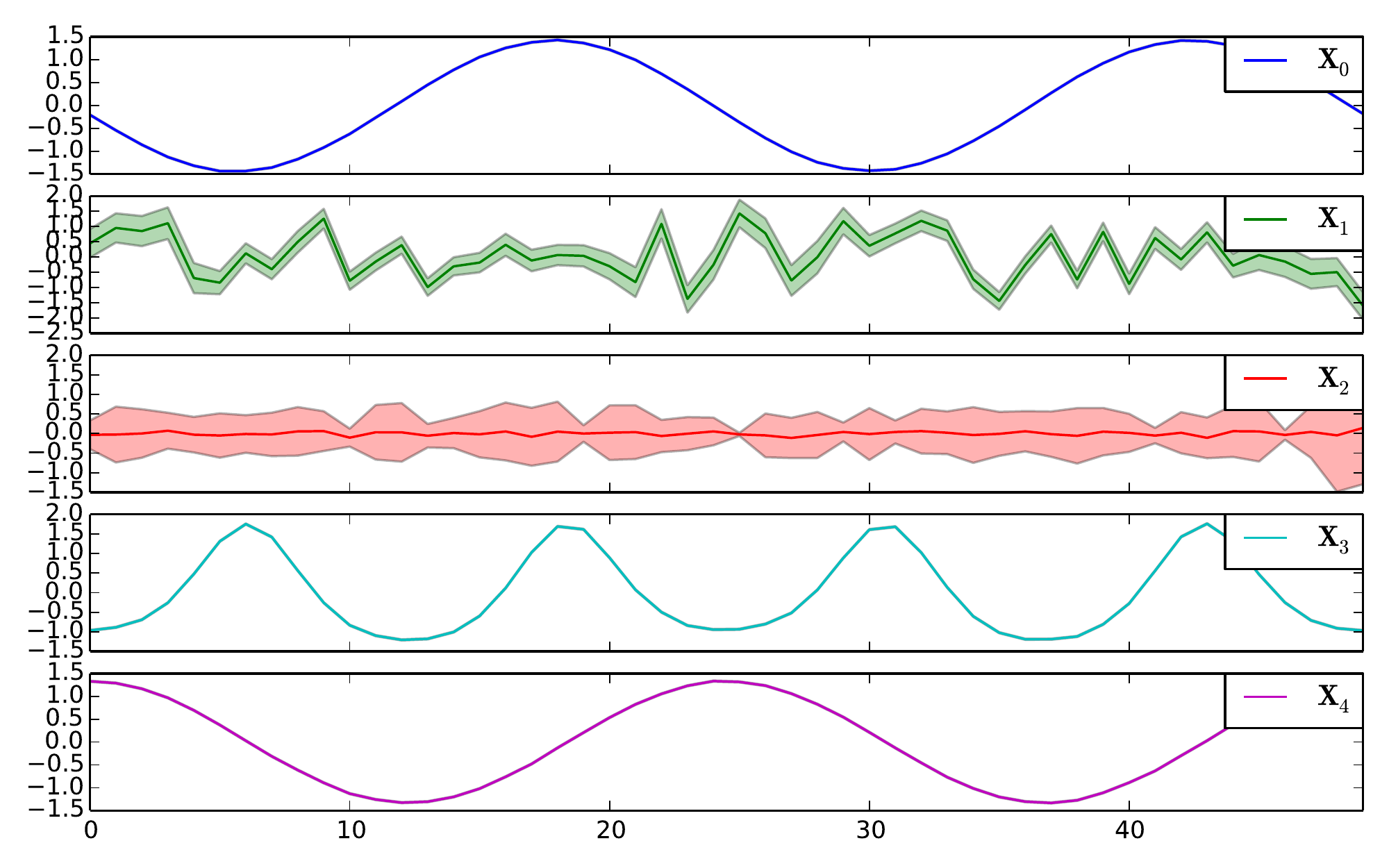}
	\label{fig:syth_ssmrdX}}\\
	\subfigure[]{\includegraphics[width=0.48\linewidth]{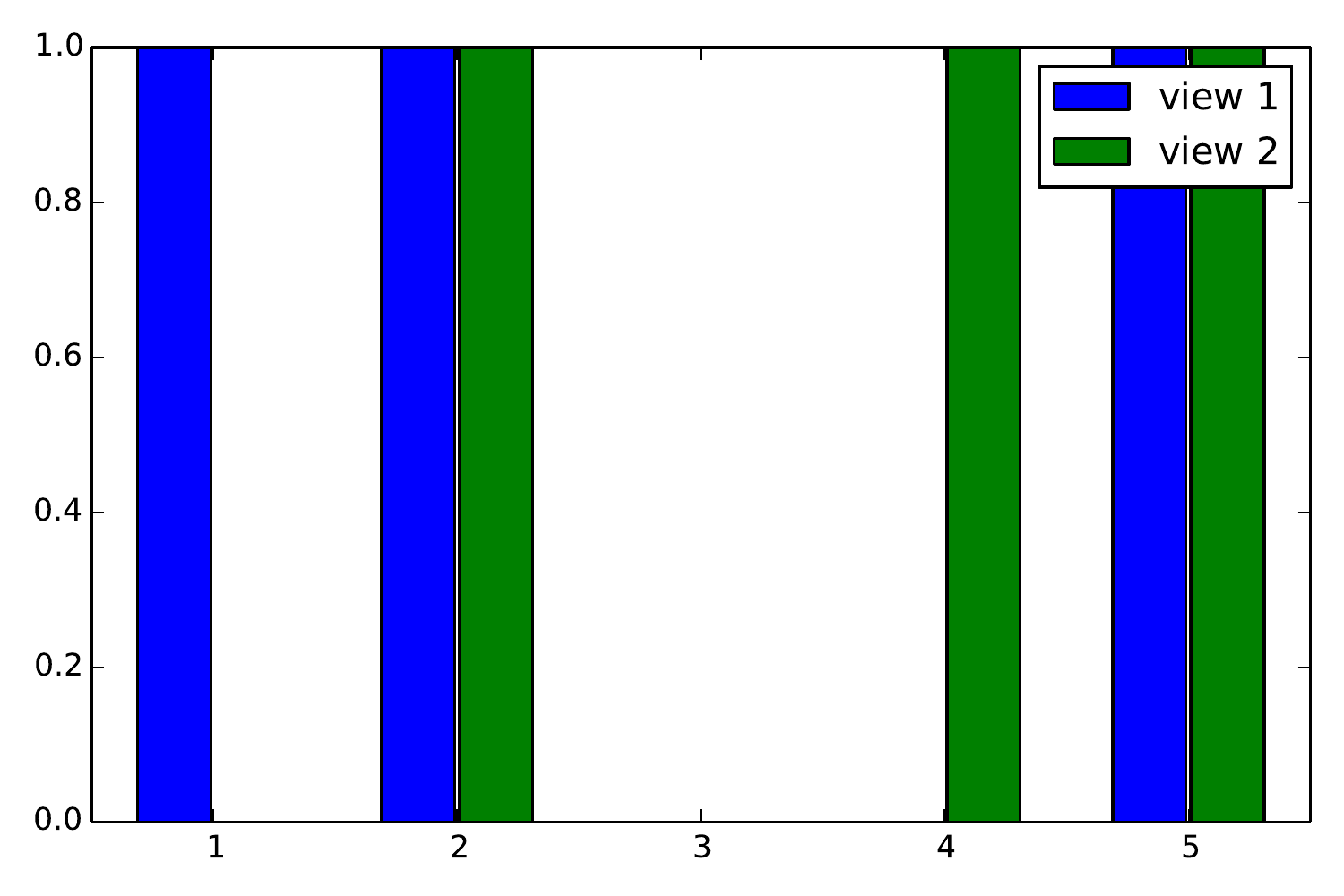}
	\label{fig:syth_ssmrdXgamma}}
	\subfigure[]{\includegraphics[width=0.48\linewidth]{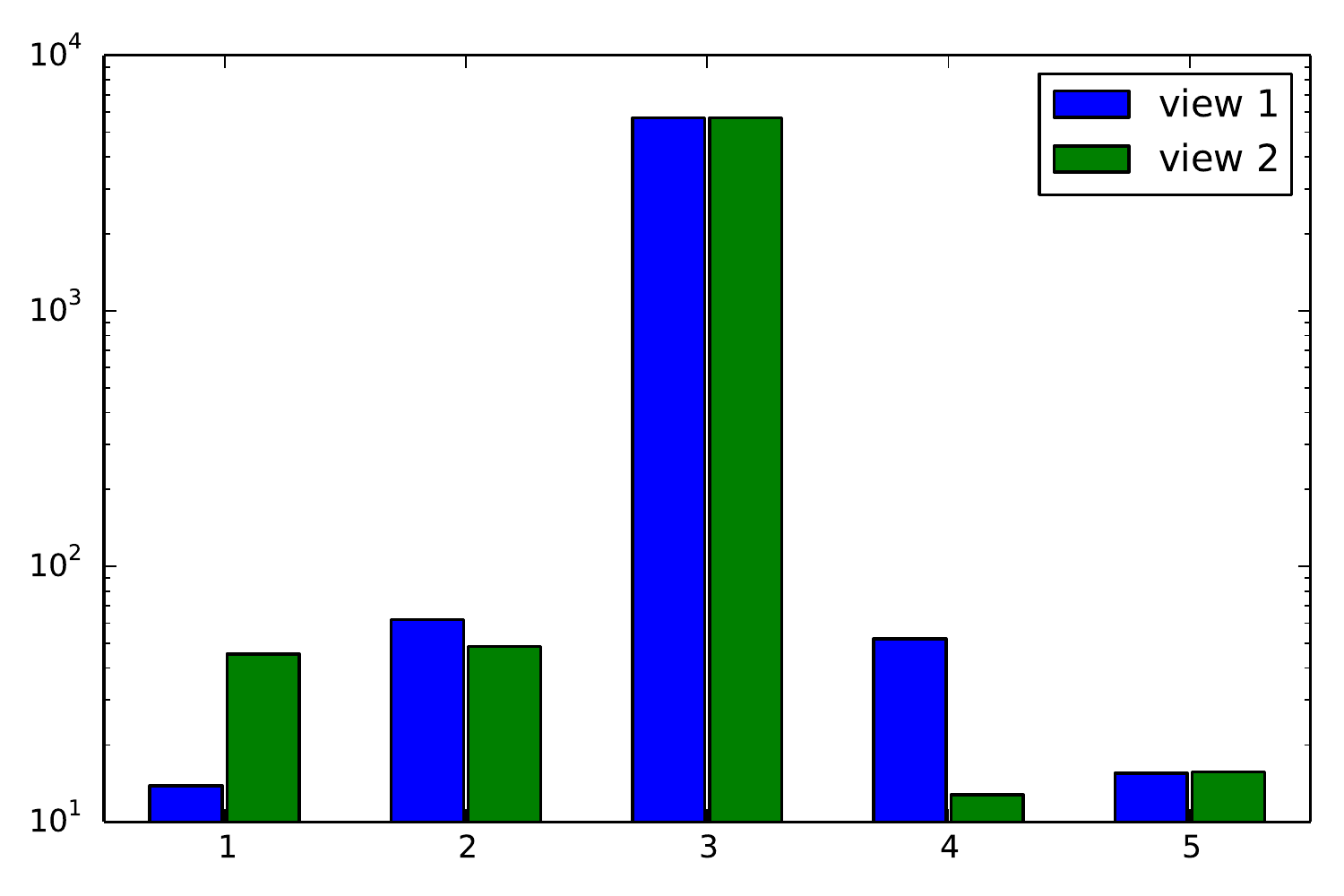}
	\label{fig:syth_ssmrdlengthscale}}
     \caption{(a)  The recovered latent signals according to our SSMRD model by assigning two views to the two observed data respectively. (b) The learned posterior probability for the switch variable $\vec{b}$. Different colors correspond to different views. (c) The learned length scales of latent dimensions.}
\end{figure}

\subsection{Synthetic Data}
We first apply both of our models to a synthetic data, where that the nature of the latent representation is known and we can ascertain when it is correctly reconstructed. We introduced three artificial signal sources, which are $y=\sin(x)$, $y=-\exp(-\cos(2x))$ and $y=\cos(x)$), and drew $50$ samples evenly from the interval between $0$ and $2\pi$ (see Fig.\ \ref{fig:syth_datasource}). The drawn samples are normalized to zero mean and unit variance. These are the latent signals that we would like to recover. We generated two sets of observed signals. For the first observed signal, we combined the $1$st and $3$rd signals, and transformed into a $12$ dimensional signals through a random linear transformation. The second observed signal are generated in the way combining the $2$nd and $3rd$ signals (see Fig.\ \ref{fig:syth_data}).

We first applied the spike and slab GP-LVM model to the first observed
signal to see whether it can recover the latent signals and determine
the used latent dimensions if you offer more latent dimensions than
the number of underlying signals. We applied our model with a linear
kernel and $5$ latent dimensions. The recovered latent signals are
shown in Fig.\ \ref{fig:syth_ssgplvmX} and
Fig.\ \ref{fig:syth_ssgplvmXgamma}. The learned $1$st and $2$nd latent
dimensions successfully recover the latent signals with very small
posterior variances, and the posterior probabilities of these two
latent dimensions are close to $1$ while the rest are close to zero,
which means the model only used the first two latent dimensions to
explain the data. The lengthscale parameters are ploted in
Fig.\ \ref{fig:syth_ssgplvmlengthscale}, where there are a big
difference between the used and unused dimensions, which matches the
observations with Bayesian GP-LVM.  It perfectly matches with the
information that we put into the data.

% mention param initialization
We then applied the spike and slab MRD model to both observed sets
signals, taking each observed signal set as a different view of the
data. The way the data was generated implies that each view has one
private latent signal and also shares a common latent signal. We aim
to recover all the latent signals with a correct assignment of latent
dimensions. We applied a linear kernel for each view and $5$ latent
dimensions. The recovered latent signals are shown in
Fig.\ \ref{fig:syth_ssmrdX} and Fig.\ \ref{fig:syth_ssmrdXgamma}. All
the three latent signals are recovered with very small posterior
variances. The $1$st view takes the $1$st latent dimension as its
private space, which recovers its private latent signal, and the $2$nd
view takes the $4$th late dimension for its private signals, where the
$5$th latent dimension are shared by both views, which recovers the
shared latent signal. The $2$nd latent dimensions are used by both
views to give some structured noise, in which the inferred variance of
the signal is significantly higher than the true signals. The inferred
length scale for the kernels of both views are shown in
Fig.\ \ref{fig:syth_ssmrdlengthscale}. Note that it is not instantly
clear how to threshold the latent dimensions according to length scale
parameters. For instance, for the $1$st view, the length scales of the
$1$st and $5$th dimensions are roughly at the same level which
corresponds to the true signals, while the $2$nd and $4$th dimensions also
have relatively low length scales. However, according to the posterior
probabilities of its switch variable, the $2$nd dimension is used by
the model while the $4$th dimension is the private space of the other
view. In this case, thresholding the length scale parameters is not
able to give the same answer as observed according to the posterior
probabilities of switch variable.

\subsection{Classification data}

\begin{figure*}[t]
        \centering
        \subfigure[]{\includegraphics[width=0.3\linewidth]{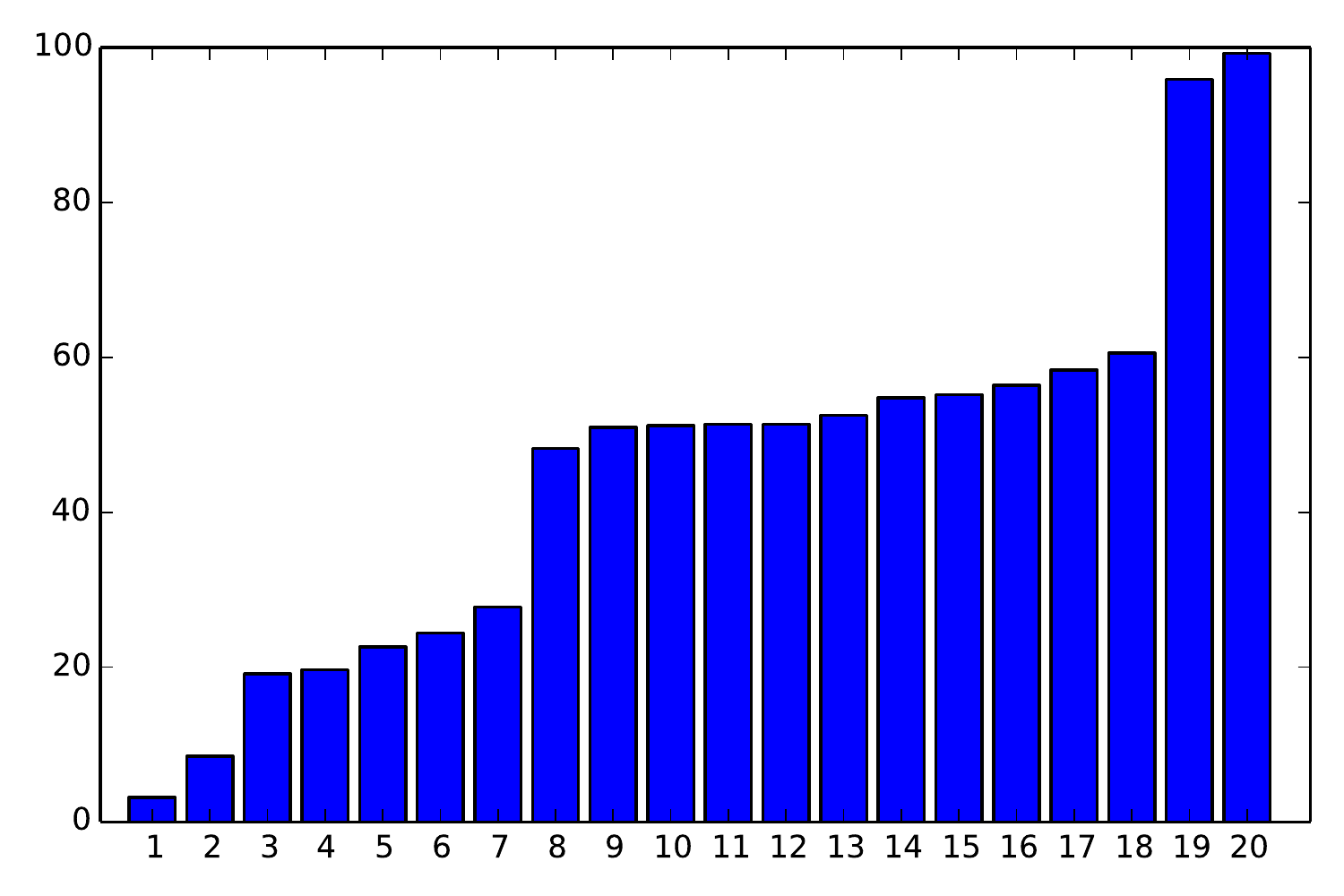}
	\label{fig:MNIST_lengthscale}}
	~\subfigure[]{\includegraphics[width=0.3\linewidth]{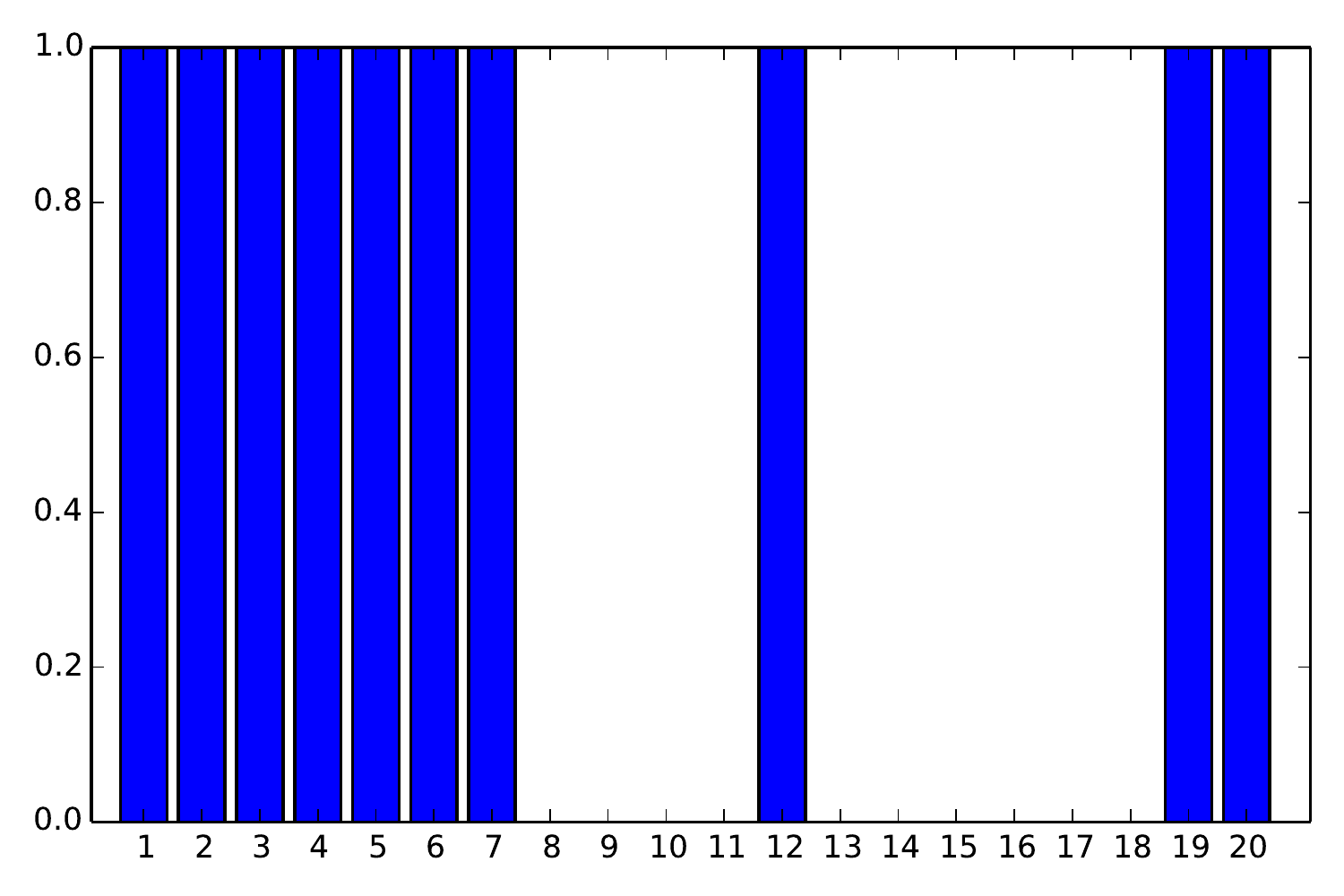}
	\label{fig:MNIST_gamma}}
	~\subfigure[]{\includegraphics[width=0.3\linewidth]{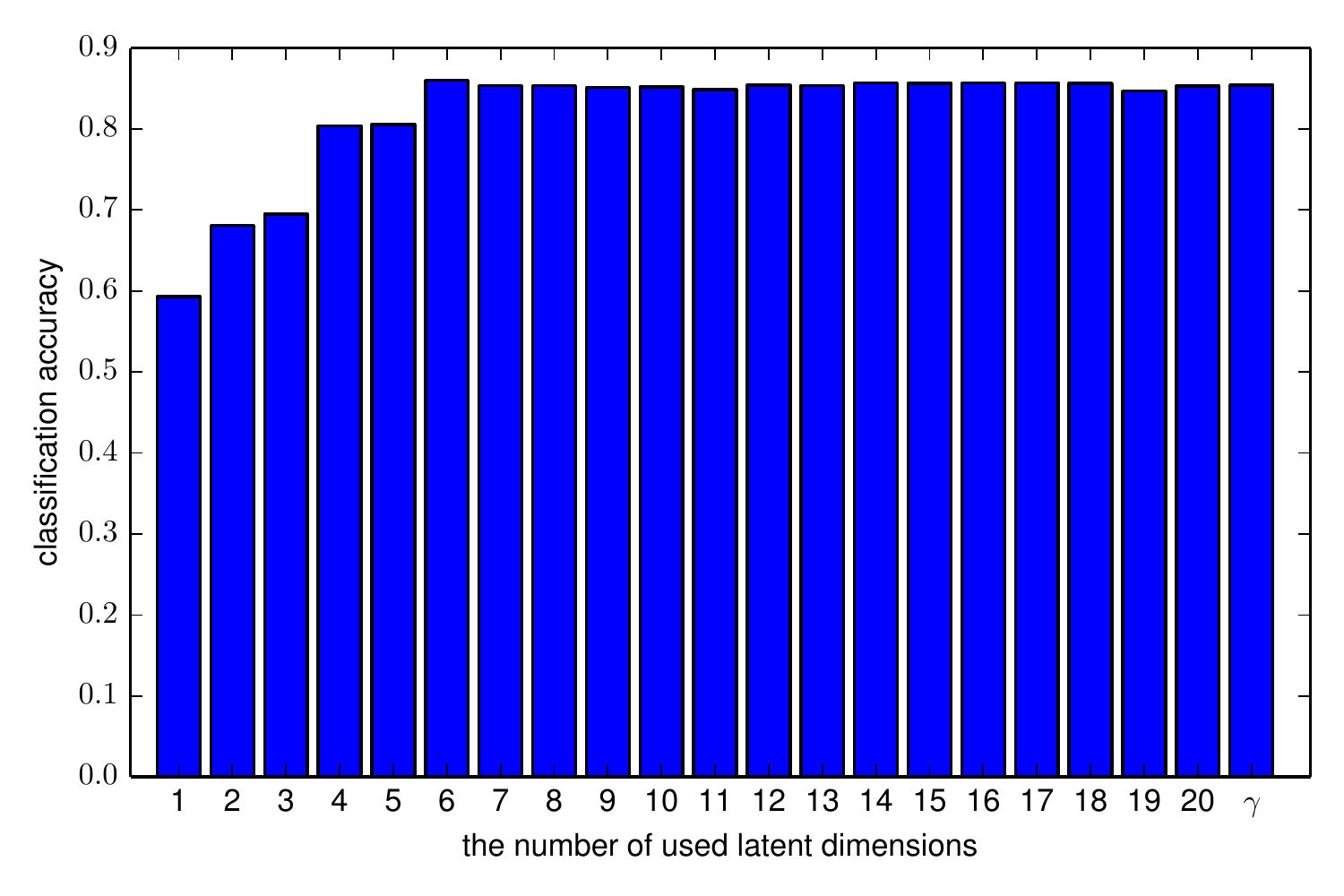}
	\label{fig:MNIST_classification}}
 \caption{(a) The learned length scales in our SSGP-LVM model from a subset of the MNIST dataset containing the digits ``1", ``7" and ``9" (sorted according to the length scale values). (b) The learned posterior probability of the switching variable $\gamma$. (c) The classification accuracy by taking different choices of latent dimensions according to their length scales, where the bar with $\gamma$ denotes the performance of choosing latent dimensions according to $\gamma$.}
\end{figure*}

We next considered a data set of hand written digits to quantify the number of latent dimensions required to represent the digits. We took a subset of images from the MNIST dataset \citep{lecun-98}. We chose the images digits of ``1", ``7" and ``9", and took 1,000 images for each character for training and 1,000 for testing.

We applied the spike and slab GP-LVM to the training set for
dimensional reduction, where the initial number of latent dimensions
was chosen to be 20. The optimization of our model parameters was done
in a purely unsupervised fashion, where no label information was
used. The resulting length scales of latent dimensions are shown in
Fig.\ \ref{fig:MNIST_lengthscale}, where the latent dimensions are
sorted according to their length scales. The posterior probabilities
of the switch variable are shown in Fig.\ \ref{fig:MNIST_gamma}. From
these values we can see that the model actively makes use of 10 latent
dimensions.

After optimizing the model parameters, we use the learned model to
infer the conditional variational posterior distributions of test
images $q_c(X_{*})$, which encode the posterior mean and variance of a
datapoint in latent space if the corresponding latent space are
used. Afterwards, we apply the nearest neighbor classifier, which
compares the posterior mean between training and testing data points,
and predict the label of a test image according to the label of its
nearest training image in the latent space. We test the classification
accuracy for different latent space configurations. We chose the
latent dimensions by thesholding their length scales at different
values, by which we obtained $20$ different choices of latent
space. We compared the performance of these choices of latent space
with choosing the latent space according to the posterior probability
of the switch variable $\gamma$. The comparison is shown in
Fig.\ \ref{fig:MNIST_classification}, where the different numbers in
$x$-axis denotes the choices corresponding to different number of used
latent dimensions, and $\gamma$ denotes the choice according to the
parameter $\gamma$.

By comparing Fig.\ \ref{fig:MNIST_lengthscale} and
Fig.\ \ref{fig:MNIST_classification}, we see that the best
classification accuracy can be obtained by using only 6 latent
dimensions, but we do not see a significant changes in length scale
between the $6$th and $7$th latent dimension. A human tentative choice
is to use 7 latent dimensions, which can give a similar level of
performance, but is difficult to automate such kind of decisions. On
the other hand, the choice according to $\gamma$ has a slightly higher
number of dimensions, but it is trivial to make automatic decision
based on that.

\subsection{Text-Image Retrieval}

An interesting application of MRD models is to relate information from
different domains. For example, relating image to text can potentially
solve ambiguities by looking at only a single view of the data. The
image representations of an object can have a lot of variances due to
changing in location, viewing angle, illumination conditions,
etc. Purely from image data, it might be difficult to figure out
different variants of the same object, but with text such ambiguities
may be resolved. Similarly, image representations can help to resolve
ambiguities in text, e.g., a facial image can easily tell different
people with the same name.

% balance the dimensionality difference
We show results on a text-image dataset collected from Wikipedia
\citep{josecp2014role}. The task here to perform multimedia
information retrieval, i.e., given a text query, the algorithm needs
to produce a ranking of the images in the training set, and similarly
given an image query to produce a ranking of texts. The Wikipedia
dataset consists of 2173/693(training/testing) image-text pairs
associated with 10 different topics. We used the features for images
and texts provided by the authors. It has a 10D text feature extracted
from a LDA model\cite{} for each document and a 128D SIFT histogram
image features for the corresponding image. The quality of the
inferred ranking is assessed in terms of mean Average Precision (mAP)
and precision-recall curves.

\begin{figure}[t]
        \centering
	\includegraphics[width=.9\linewidth]{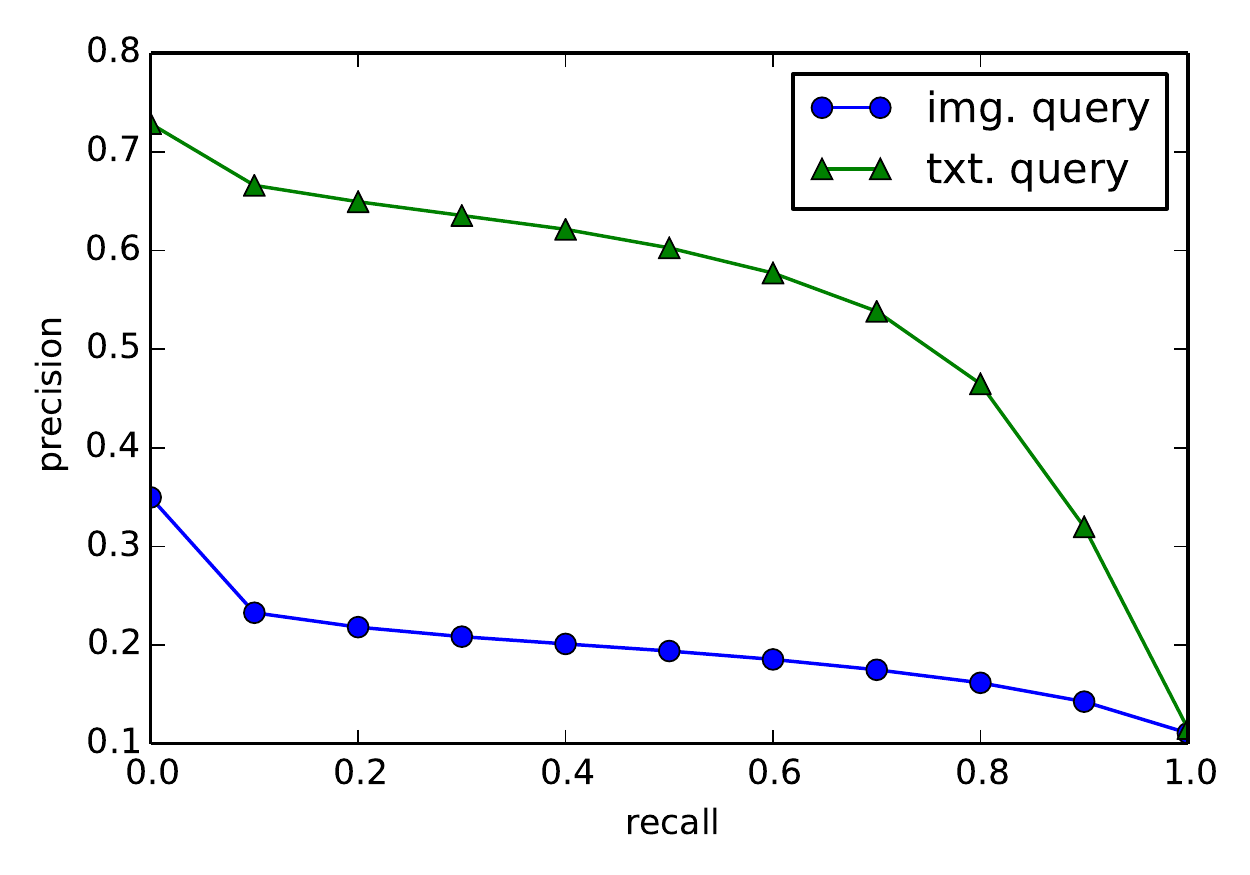}
     \caption{The precision-recall curve for both image and text queries on the Wiki dataset.}\label{fig:wiki_PRcurve}
\end{figure}

We applied our SSMRD model by assigning image features to a view and
text features to another view. Both image and text features are
normalized to zero mean and unit variance before inference. To
overcome the unbalanced dimensionality between image and text features
(128 v.s.\ 10), we replicate the text features 6 times, making it a
60D representation. An exponentiated quadratic kernel is used for each
view and the number of latent dimensions is chosen to be 10. We
initialize the mean of variational posterior $\mu$ according to the
topic of each image-text pair by placing them on the vertices of a
simplex structure. The rest of learning runs without taking into
account the topic information. The model learns a shared latent space
between images and texts as well as their private latent space. After
optimizing all the model parameters, the query results are produced by
first searching the conditional variational posterior distribution
$q(X_{*}| \vec{b}^{(c)})$ given the query input, i.e., image or text
features $Y^{(c)}_{*}$, and ranking the training data according to the
euclidean distance in the shared latent space. The qualities of the
produced rankings are evaluated in terms of precision-recall curves
(see Fig.\ \ref{fig:wiki_PRcurve}) and mAP (see
Tab.\ \ref{tab:wiki_map}).

\begin{table}[t]
\centering
\caption{Mean Average Precision (mAP) Scores} \label{tab:wiki_map}
\begin{tabular}{|l|l|l|l|l}
\cline{1-4}
      & img. query & txt. query & avg.  &  \\ \cline{1-4}
SSMRD & 0.170      & \textbf{0.540}      & \textbf{0.355} &  \\ \cline{1-4}
SCM   & \textbf{0.362}      & 0.273      & 0.318 &  \\ 
SM    & 0.350      & 0.249      & 0.300 &  \\ 
CM    & 0.267      & 0.219      & 0.243 &  \\ \cline{1-4}
GMMFA & 0.264      & 0.231      & 0.248 &  \\ 
GMLDA & 0.272      & 0.232      & 0.253 &  \\ \cline{1-4}
\end{tabular}
\end{table}

The performances of state of art algorithms are also shown in
Tab.\ \ref{tab:wiki_map}. All the results used the same feature set
extracted with dataset. Correlation matching (CM), semantic matching
(SM), and semantic correlation matching (SCM) are the methods proposed
by the creators of the dataset \citep{josecp2014role}, of which SCM
gives the state of art performance. The results with generalized
multiview analysis (GMA) with LDA (GMLDA) and marginal Fisher analysis
(GMMFA) are reported by \cite{SharmaEtAl2012}. The mAP measures are
directly taken from their papers. Our performance for text queries is
significantly better than all the state of art algorithms. Its
precision-recall curve drops much slower, compared with what is
reported in \cite{josecp2014role}. Our performance for image queries
is below the state of art performance. We suspect it is due to lack of
enough inducing inputs (100 is used for the reported performance),
which directly limits the modeling capability, and not introducing
label information into learning. Note that all their algorithms except
$CM$ are supervised algorithms, while our model does not make use of
label information during training except initializing the latent
space.

\section{Conclusion}

Standard approaches to variable selection in Gaussian process latent
variable models have relied on scaling priors to reduce the influence
of particular dimensions. We have introduced switching variable and a
\emph{spike and slab} prior which allows us to explicitly model the
switching on and off of particular latent dimensions. This provides a
more principled approach for selecting latent dimensions. By
variationally integrating out the spike and slab latent variable we
derived a lower bound on the log marginal likelihood. For efficient
implementation we used a parallel version of the algorithm with GPU
acceleration. In the GP-LVM, multiple view learning is achieved
through \emph{manifold relevance determination}, where the choices of
latent dimensions for different views are explicitly modelled. We also
applied the spike and slab approach to the MRD prior and were able to
show significantly better than state of the art performance on a
cross-modal multimedia retrieval task.

Structural learning in Gaussian process models is becoming more
important with the advent of deep Gaussian processes \cite{DamianouLawrence2013}. We
envisage that the combination of spike and slab models, alongside
appropriate infinite binary process (IBP) priors \cite{GriffithsGhahramani2005} will enable
structural learning of the composite process models.

%\subsubsection*{Acknowledgements}
%
%Use unnumbered third level headings for the acknowledgements.  All
%acknowledgements go at the end of the paper.  Be sure to omit any
%identifying information in the initial double-blind submission!
%
% Jorg Lucke

{
%\small
%\bibliographystyle{abbrvnat} %abbrv
\bibliographystyle{icml2015}
\bibliography{spike_and_slab,../../../bib/lawrence,../../../bib/other,../../../bib/zbooks}
}
\end{document}